\def\year{2019}\relax
\documentclass[letterpaper]{article} 
\usepackage{aaai19}  
\usepackage{times}  
\usepackage{helvet}  
\usepackage{courier}  
\usepackage{url}  
\usepackage{graphicx}  
\usepackage{amsmath}
\usepackage{amssymb}
\DeclareMathOperator*{\argmax}{argmax}
\usepackage{floatrow}  
\newfloatcommand{capbtabbox}{table}[][\FBwidth]  
\title{Turbo Learning Framework for Human-Object Interactions Recognition and Human Pose Estimation}
\author{Wei Feng\textsuperscript{\rm 1}, 
	Wentao Liu\textsuperscript{\rm 2, 1},
	Tong Li\textsuperscript{\rm 1}, 
	Jing Peng\textsuperscript{\rm 1},
    Chen Qian\textsuperscript{\rm 1},
    Xiaolin Hu\textsuperscript{\rm 2}\\
	\textsuperscript{\rm 1} SenseTime Group Ltd.\\
	\textsuperscript{\rm 2} Department of Computer Science and Technology, Tsinghua University\\
	\{fengwei, litong, pengjing1, qianchen\}@sensetime.com,  liuwtwinter@gmail.com, xlhu@tsinghua.edu.cn
}
\begin{document}
\maketitle
\begin{abstract}
Human-object interactions (HOI) recognition and pose estimation are two closely related tasks. Human pose is an essential cue for recognizing actions and localizing the interacted objects. Meanwhile, human action and their interacted objects' localizations provide guidance for pose estimation. In this paper, we propose a turbo learning framework to perform HOI recognition and pose estimation simultaneously. First, two modules are designed to enforce message passing between the tasks, i.e. pose aware HOI recognition module and HOI guided pose estimation module. Then, these two modules form a closed loop to utilize the complementary information iteratively, which can be trained in an end-to-end manner. The proposed method achieves the state-of-the-art performance on two public benchmarks including Verbs in COCO (V-COCO) and HICO-DET datasets. 
\end{abstract}

\section{Introduction}

Human-object interactions (HOI) recognition~\cite{gkioxari2017detecting,gupta2009observing,yao2010modeling,chen2014predicting} aims to detect and recognize triplets in the form $<human,action,object>$ from a single image. It has attracted increasing attention, due to its wide applications in image understanding and human-computer interfaces. Although previous methods have made significant progress, it is still a challenging task due to the diversity of human activities and the complexity of backgrounds.

Most existing methods perform HOI recognition in two steps. First, a detector is employed to detect all objects in the image. Then the action and interacted objects' locations are obtained mainly based on the human appearance. However, these methods are easily influenced by the change of human appearance. In contrast to source human appearance, human pose contains structure information, which is a more robust reference for both action recognition and object localization. Though pose information also comes from appearance features, from the viewpoint of HOI recognition task, it provides an indirect way to encode appearance features. Therefore, considering pose estimation in HOI recognition is thus a natural and intuitive idea. \citeauthor{gupta2015visual}~(\citeyear{gupta2015visual}) propose a dataset named Verbs in COCO (V-COCO), in which two kinds of information are labeled at the same time. However, most existing methods treat HOI recognition and pose estimation as separate tasks, which ignore the information reciprocity between two tasks. 

\begin{figure*}
	\centering
	\begin{minipage}[t]{1\linewidth}
		\centerline{\includegraphics[width=0.8\textwidth]{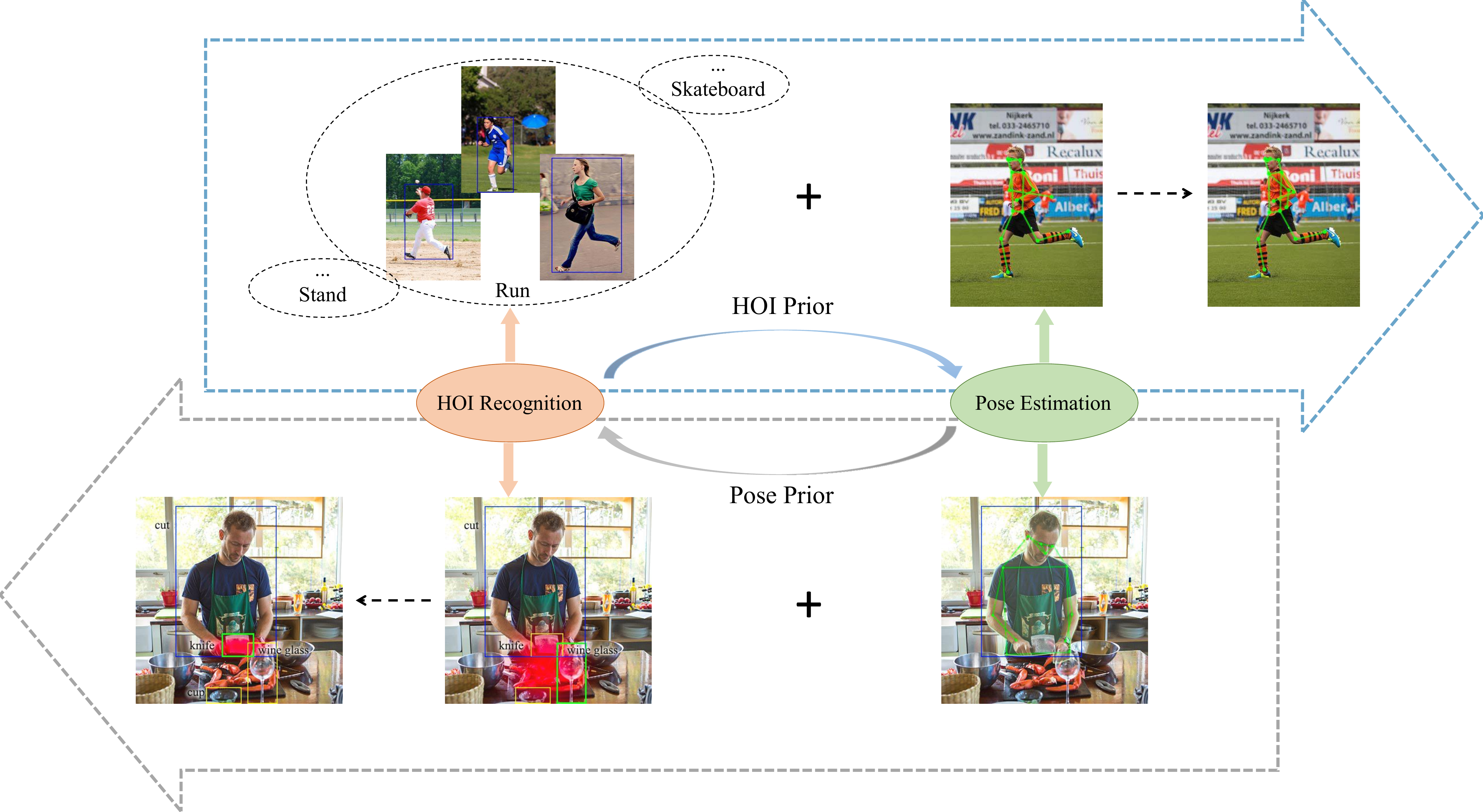}}
	\end{minipage}
	\caption{HOI recognition and pose estimation can help each other. People doing the same action like running have similar keypoints distribution. When there is a priori action pattern, the wrong pose estimation might be corrected.
		Meanwhile, when there are several objects in the estimated action-type specific density area (red area), the model will be confused to find the target object (green box). However, Some keypoints like wrist can refine the red area and help the model to localize the interacted object successfully.}
	\label{fig:introduction}
\end{figure*}

Constructing a closed loop between HOI recognition and pose estimation may bring improvement to each task as shown in Fig~\ref{fig:introduction}. On one hand, human pose can improve the robustness of HOI recognition. As for action recognition, human pose offers accurate analysis of human structure rather than coarse information of human body appearance, thus reducing the impact of changes in human appearance. As for target localization, the locations of some keypoints can reveal the location and direction of the interacted objects. For example, the keypoint of the wrist is very close to the object in the action $hold$. Therefore, human pose is a powerful cue for recognizing actions and localizing the interacted objects. On the other hand, human action can guide the distribution of keypoints vice versa. As a result, the relative position between the keypoints is clearer. Meanwhile, their interacted objects' localization can help localize specific keypoints. So human action and their interacted objects' localization provide guidance for pose estimation.

In this paper, we propose a turbo learning method to simultaneously perform HOI recognition and pose estimation. Specifically, a pose aware HOI recognition module is used to perform HOI recognition based on both image and pose features, which can reduce the influence of changes in human appearance. Meanwhile, a HOI guided pose estimation module is used to perform pose estimation with clear relative distribution between keypoints, in which HOI features are encoded into space location information. Then these two modules form a closed loop in the similar way as an engine turbo-charger, which feeds the output back to the input to reuse the exhaust gas for better engine efficiency. The feedback process can gradually improve the results of both tasks. To the best of our knowledge, this is the first unified end-to-end trainable network for simultaneous HOI recognition and pose estimation.

Our contributions are in three folds: (1) A pose aware HOI recognition module is proposed, in which the human pose can help to extract more accurate human structure information than the source appearance features. (2) A HOI guided pose estimation module is introduced, where HOI recognition features and image features form an attention mask for pose estimation, which transforms the pattern of human action into the general distribution of keypoints. (3) A turbo learning framework is proposed to sequentially perform HOI recognition and pose estimation using the two modules, which can gradually improve the results on both tasks. The proposed method achieves the state-of-the-art results on two public benchmarks including V-COCO and HICO-DET~\cite{chao2017learning}.

\section{Related Work}

\subsection{HOI Recognition}

HOI recognition is not a new problem in computer vision~\cite{herath2017going}. Traditional methods~\cite{Hu2014Recognising,Delaitre2010Recognizing,pishchulin2014fine} usually use many contextual elements in images to predict action categories, including human pose, manipulated objects, scene, and other people in images~\cite{gupta2009observing,maji2011Action,Desai2012Detecting}. Likewise, deep methods also use one or several of these elements to recognize HOIs, but most of them only consider human appearance to be the key point of HOI recognition. For example, \citeauthor{mallya2016learning}~(\citeyear{mallya2016learning}) fuse CNN-based human appearance features and global context features to achieve the state-of-the-art performance on predicting HOI labels; \citeauthor{gkioxari2017detecting}~(\citeyear{gkioxari2017detecting}) use merely human features to achieve the state-of-the-art performance in HOI recognition. Beyond that, \citeauthor{chao2017learning}~(\citeyear{chao2017learning}) use spatial relations between human and object positions to recognize HOIs. \citeauthor{shen2018scaling}~(\citeyear{shen2018scaling}) focus on the difficulty of obtaining all the possible HOI samples in reality, and propose a zero-shot learning method to tackle with the lack of data problem.

\subsection{Combination of Action Recognition and Pose Estimation}
It's not difficult to understand that there exists an intrinsic connection between human actions and human poses~\cite{wei2016convolutional,liu2018cascaded,cao2017realtime,ning2017knowledge,luvizon2017human,xiaohan2015joint}. Different people may have different skin colors and appearance, dressing various clothes, but their poses are similar when they are doing the same action due to the homogeneity of human body. Intuitively, adding pose information would be beneficial for deep networks to recognize HOI categories, but relevant researches are scarce, especially in deep learning. In early times, \citeauthor{yao2010modeling}~(\citeyear{yao2010modeling}) use a random field model to encode mutual context of human pose and objects, but their method needs to generate a set of models, each modeling one type of human pose in one action class. \citeauthor{Desai2012Detecting}~(\citeyear{Desai2012Detecting}) propose a compositional model that uses human pose and interacting objects to predict human actions, but the visual phraselets and tree structure they use are too simple to capture sophisticated HOI relations in large datasets. In connection with neural networks, \citeauthor{shen2018scaling}~(\citeyear{shen2018scaling}) concatenate pre-computed pose features and appearance features to improve model performance on HOI recognition. Recently, \citeauthor{luvizon20182d}~(\citeyear{luvizon20182d}) design a single architecture for jointly 2D and 3D pose estimation from still images and action recognition from video sequences. Similar to their idea, our method combines two tasks using one end-to-end trainable framework. On top of that, we adopt a novel turbo learning framework that leads to better results in each sub-tasks.

\begin{figure*}
	\centering
	\begin{minipage}[t]{1\linewidth}
		\centerline{\includegraphics[width=1\textwidth]{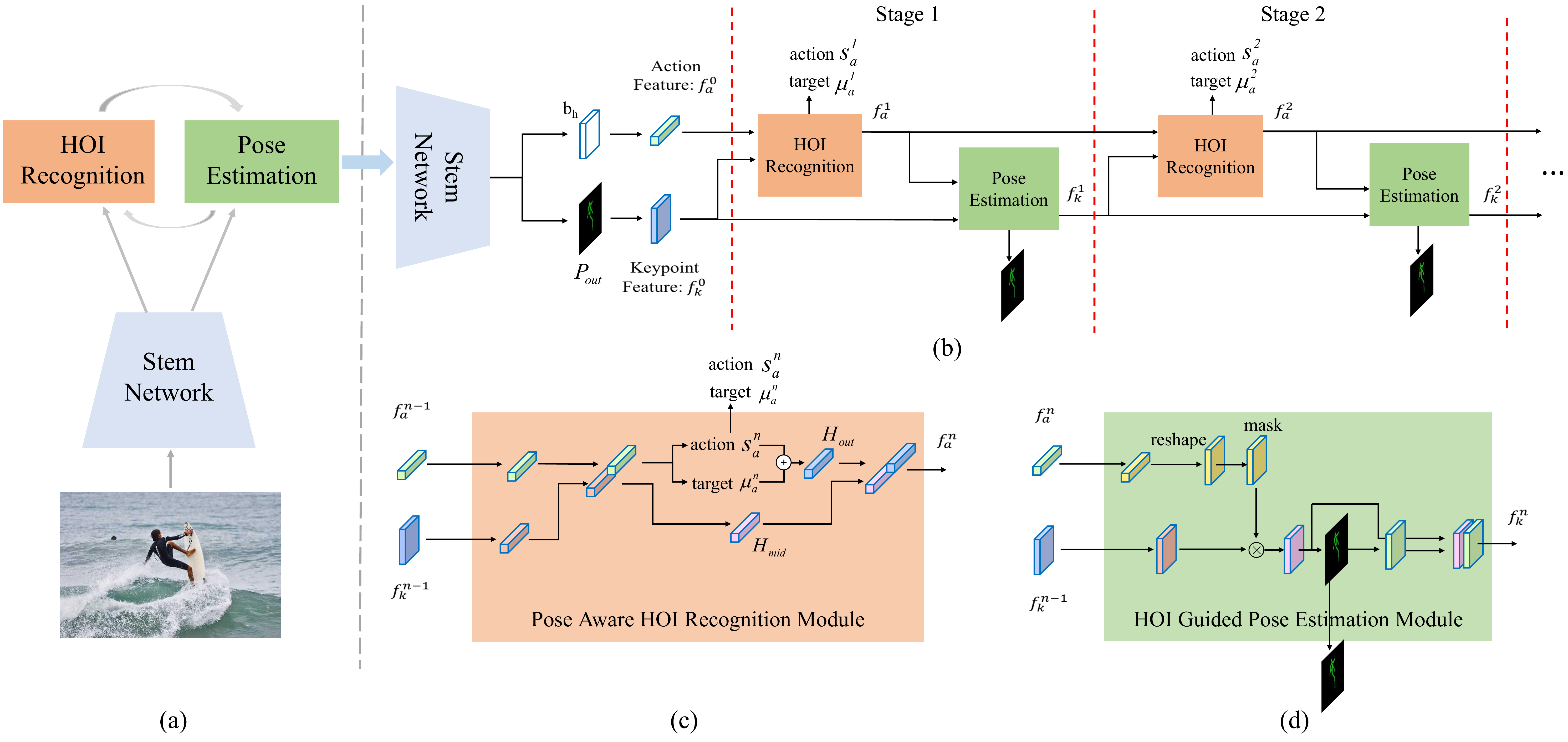}}
	\end{minipage}
	\caption{(a) Overview of the proposed method. (b) Unfolded diagram of the turbo learning framework along time. (c) The architecture of pose aware HOI recognition module. (d) The architecture of HOI guided pose estimation module.}
	\label{fig:framework}
\end{figure*}

\section{Method}

An overview of our framework is illustrated in Fig~\ref{fig:framework}(a). We propose an end-to-end turbo learning framework to integrate two different tasks. After extracting the human appearance features by the stem network, the pose aware HOI recognition module is applied to predict HOI with both human appearance features and human pose features. Then, the proposed HOI guided pose estimation module updates the human pose result depending on HOI recognition result. The two processes repeat for several times. The two modules form  a closed loop to gradually improve the results of both pose estimation and HOI recognition. The closed loop can be expanded by the time sequence as shown in Fig~\ref{fig:framework}(b).


\subsection{Pose Aware HOI Recognition}

Inspired by~\cite{gkioxari2017detecting}, we decompose HOI recognition into two subtasks: action recognition and target localization. We treat the action recognition task as a multi-label task, since a person can simultaneously perform multiple actions (\emph{e.g.}, $sit$ and $hold$). To avoid competition between classes, a set of binary sigmoid classifiers are used for multi-label action classification as in~\cite{gkioxari2017detecting}. Hence, the action recognition model outputs a confidence score $s_a$ for action $a$.   Normally, the action score $s_a$ is calculated only based on the human appearance features $F$, which can be written as $s_{a} = sigmoid(I(F))$, where $I(\;)$ refers to the fully connected transformation of the feature maps.

Previous methods predict two subtasks with appearance features extracted from human bounding box. However, we notice that the human region contains lots of background information and the rough feature of human make the HOI recognition easily influenced by the change in appearance. Instead of only considering the human appearance feature, we argue that the human pose estimation obtains detailed analysis of human structure and could provide robust pose prior for HOI recognition.

Therefore, we design two types of pose aware HOI modules to integrate human pose constraints. The first one is a simple multi-task training, in which we extend the HOI recognition branch with a pose estimation branch. It needs to be noticed that these two tasks only share stem block. In the second design, instead of integrating human pose information implicitly, we further encode the human pose features as input of HOI recognition task as shown in Fig~\ref{fig:framework}(c). In order to make full use of pose information, the pose estimation features $f_{k}$ consist of two parts: the intermediate features $P_{mid}$ from the eighth convolution layer of pose estimation branch, and the pose estimation branch's output $P_{out}$. The feature $P_{mid}$ contains rich information of human pose and it is abstract enough to encode all body part locations. The estimation result $P_{out}$ provides the exact structure of the human body. 
Furthermore, the HOI features in the previous stage also contribute to the HOI recognition in this stage, which will be discussed in the later. Therefore, the concatenation of pose estimation features, human appearance features $F$ and the HOI features in the previous stage are used as input $h$ of this module, which can be written as:
\begin{equation}
h = I([f_{k}^{n-1},F,H_{mid}^{n-1},H_{out}^{n-1}]),
\end{equation}
where $[\;]$ refers to the concatenation of the feature maps.  $H_{mid}^{n-1}$ and $H_{out}^{n-1}$ represent the intermediate features and recognition results of the previous HOI recognition module respectively.

For the target localization task, each action predicts the location of the associated object. To enforce the prior that the interacting objects are around the human body, the position of each object is represented relative to the human proposal as $\hat{b_{o|h}}$. 
\begin{equation}
\hat{b_{o|h}} = \{\frac{x_o - x_h}{w_h}, \frac{y_o - y_h}{h_h}, \log\frac{w_o}{w_h}, \log\frac{h_o}{h_h}\},
\label{encoding}  
\end{equation}
where $h_o$, $w_o$ indicate the height and width of the ground truth target object, and $h_h$, $w_h$ denote the height and width of the human region proposal (\emph{i.e.}, its IoU overlap with the ground-truth box is $\ge$ 0.5.). However, the human region proposal and target object
may vary in sizes, so we use the smooth $L_1$ loss as it is less sensitive to outliers. Note that the actions without interactive objects will not produce loss in this task.

Predicting the precise location is a challenging task. For example, the target object usually does not appear in the region proposal $b_h$ for the action $throw$. Therefore, in the test phase, the target localization branch estimates a probability of object location instead of the precise position.  
We combine the predicted probability $\mu_a$ and the detected objects position $b_{o|h}$ from the object detection branch to precisely localize the target. Specifically, given the human box $b_h$ and action $a$, the target localization branch outputs the relative position of associated object $\mu_a$, then the target localization probability $g_{h,o}^a$ can be written as:
\begin{equation}
g_{h,o}^a = \exp(-||b_{o|h} - \mu_a||^{2}/2\sigma^2),
\end{equation}
where $\sigma$ is a hyperparameter to constrain search region, we set it to 0.3 in the experiment. The detected object position $b_{o|h}$ is also encoded as Equation \ref{encoding}.
After that, the accurate localization of object $\hat{b_{o|h}}$ is obtained by $\hat{b_{o|h}}= \argmax_{b_{o|h}}{g_{h,o}^a}$.

\subsection{HOI Guided Pose Estimation}
The previous module has predicted action confidence and the position of the associated object, then we propose a HOI guided pose estimation module to improve the prediction of human pose. 
However, the results of action recognition and target localization are predicted by a fully connected network and lots of spatial structures are lost. To encode HOI feature into spatial constraint, which is important to human pose estimation, we 
propose a HOI attention mask to refine keypoint detection.

The architecture of HOI guided pose estimation module is shown in Fig~\ref{fig:framework}(d). In this module, an attention mask is generated by the HOI features and then the attention mask is performed on the keypoint feature. The modulated features are utilized to generate refined results of human pose estimation. The HOI features $f_{a}$ include intermediate features $H_{mid}$ and recognition output $H_{out}$ simultaneously, where $H_{out}$ consists of the results of action recognition and target localization, and $H_{mid}$ contains the implicit information about human action patterns.
More formally, we define the attention mask $Att_{action}$ as Equation \ref{mask}, and use $R(\;)$ to indicate reshape operation.
\begin{equation}
\label{mask}
Att_{action} = R(sigmoid(I([f_{a}]))),
\end{equation}

After that, the modulated keypoint features $p$ are achieved as the element-wise multiplication results of attention mask and pose estimation features in previous stage $f_{k}^{n-1}$. $f_{k}^{n-1}$ denotes the features  of the previous pose estimation module. Specifically, given attention mask $Att_{action}$, we generate the input feature maps $p$ as follows:
\begin{equation}
p = Att_{action} \cdot f_{k}^{n-1}.
\end{equation}

Then we feed the modulated features $p$ into the keypoint detection block to predict the human pose estimation result in stage n. After the attention mask filtering, some false located keypoints are corrected as shown in experiments.
The keypoint detection block consists of sequential 3$\times$3 512-d convolution layers, one deconv layer, one upsample layer and the final convolution layer which projects 512 channels of feature maps into K masks (K is the number of keypoints). We model the location of ground truth as a one-hot mask like~\cite{He2017Mask}. Specifically, the ground truth mask is a one-hot $m\times m$ binary mask where only the pixel at the ground truth location is foreground. Therefore, we regard the pose estimation task as a kind of m$\times$m categories of classification tasks, so the training objective is to minimize the cross-entropy loss over an $m\times m$-way softmax output. Similar as the HOI recognition module, the pose estimation module only calculate the loss of region proposals whose overlap of ground truth human box $b_h$ exceed overlap threshold. Meanwhile, the invisible keypoints will not cause loss.


\subsection{Turbo Learning Framework}
\label{task}
Better pose estimation results lead to better HOI recognition result and better HOI recognition results lead to more robust pose estimation.
To utilize the flow of complementary information iteratively, we introduce a turbo learning framework, which can be expanded as a sequence of pose aware HOI recognition modules and HOI guided pose estimation modules by the time step. Among them, a pose aware HOI recognition module and a HOI guided pose estimation module form a stage as shown in Fig~\ref{fig:framework}(b). On the one hand, the pose features provide the subsequent HOI recognition module an expressive non-parametric encoding of each keypoint, allowing the HOI recognition module to learn explicit human appearance and location of the keypoints that need to be focused on. On the other hand, the HOI features provide the subsequent pose estimation module a comprehensive encoding of a action pattern from all action classes, and the general scope of some keypoints.
As a result, each stage of the turbo learning framework increasingly refines the action classification results, the locations of target objects and each keypoint. This framework is fully differentiable and therefore can be end-to-end trainable using backpropagation. The whole loss function can be written as:
\begin{equation}
L = \sum_{i=1}^{N}(\lambda_{pose}^{i}L_{pose}^{i} + \lambda_{HOI}^{i}L_{HOI}^{i}) + L_{det},
\end{equation}
where $L_{det}$ is the loss function for the detection task, $L_{HOI}^{i}$ and $L_{pose}^{i}$ are losses for HOI recognition and pose estimation in stage $i$. $\lambda_{pose}^{i}$ and $\lambda_{HOI}^{i}$ are the hyper-parameters to control the balance in stage $i$. N is the total number of stages.

In each stage of the turbo learning framework, the input consists of three parts: human appearance features $F$, HOI features $f_{a}^{n-1}$ and the pose features $f_{k}^{n-1}$. Among them, the latter two parts are produced in the previous stage. 
Specifically, the set of input feature maps $t^{n}$ in stage $(n < N - 1)$ can be written as:
\begin{equation}
t^{n} = \left \{ F, f_{k}^{n-1}, f_{a}^{n-1} \right \},
\end{equation}
where $f_{k}^{n-1}$ and $f_{a}^{n-1}$ denote the pose features and HOI features in the $(n-1)$-th stage respectively. 
The $n$-th stage not only outputs the recognition results $[s_{a}^{n}, \mu_a^n]$ and human pose heatmap $k^n$, but also the feature maps for the $(n + 1)$-th stage, which contain pose features $f_{k}^{n}$ and HOI features $f_{a}^{n}$.

\begin{figure*}
	\centering
	\begin{minipage}[t]{1\linewidth}
		\centerline{\includegraphics[width=1\textwidth]{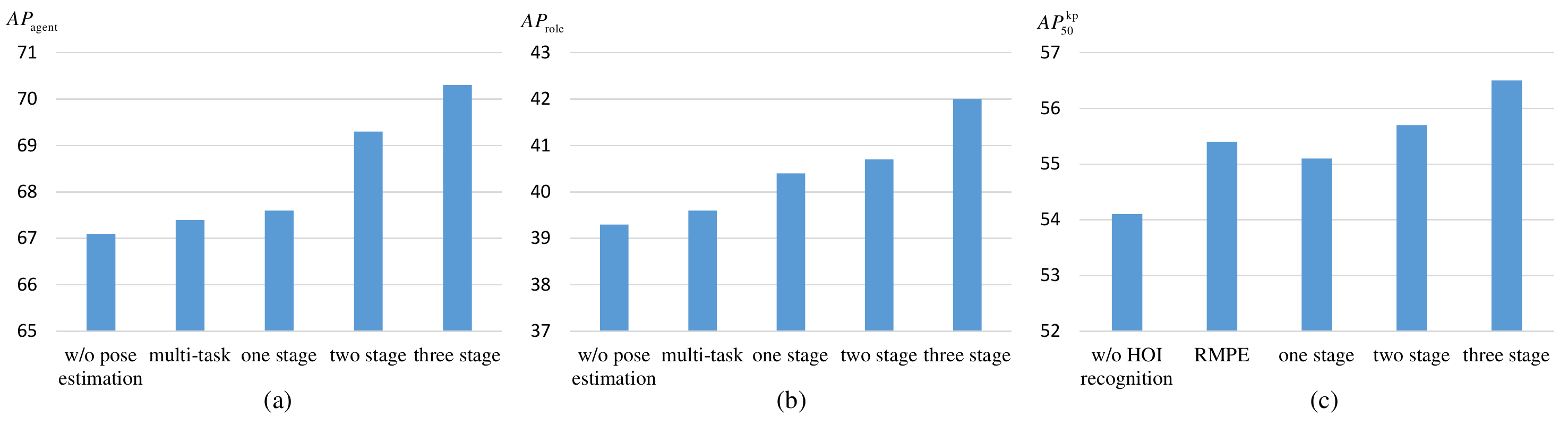}}
	\end{minipage}
	\caption{Recognition results on the V-COCO test set. From left to right: $AP_{agent}$, $AP_{role}$ for HOI recognition and $AP_{50}^{kp}$ for pose estimation.}
	\label{fig:HOI}
\end{figure*}

\section{Experiment}

\subsection{Datasets and Metrics}

\textbf{V-COCO:} V-COCO is a subset of COCO~\cite{lin2014microsoft}, which is commonly used for HOI recognition. This dataset includes $\sim$5k images in the trainval set and $\sim$5k images in the test set. It annotates 17 keypoints of the human body and 26 common action classes. Three actions (cut, hit, eat) are annotated with two types of targets: instrument and direct object, which means that the action is associated with two objects. For example, one man cuts the pizza with a knife. As accuracy is evaluated separately for the two types of targets, we extend the 26 action classes to 29 classes, same as in~\cite{gkioxari2017detecting}. For evaluation, we adopt the two Average Precision (AP) metrics as in~\cite{gupta2015visual}. The 'agent AP' (AP$_{agent}$) evaluates the AP of the pair $<human, action>$. Note that AP$_{agent}$ does note require localizing the target, so we pay more attention to AP$_{role}$ which evaluates the AP of the triplet $<human, verb, object>$.

\textbf{HICO-DET:} the HICO-DET dataset is an extension to the ``Humans Interacting with Common Objects" (HICO) dataset~\cite{chao2015hico}. In HICO-DET, the annotation of each HOI includes a human bounding box and an object bounding box with a class label respectively. HICO-DET contains about 48k images, $\sim$38k for training and $\sim$9k for testing. It includes 600 interaction types, 80 unique object types that are identical to the COCO categories, and 117 unique verbs. The official evaluation code of HICO-DET reports the mean AP over Full test set (including all 600 HOI categories), Rare test set (including HOIs with less than 10 training instances only) and Non-Rare test set (including HOIs with 10 or more training instances only) respectively.

\begin{figure*}
	\centering
	\begin{minipage}[t]{1\linewidth}
		\centerline{\includegraphics[width=0.8\textwidth]{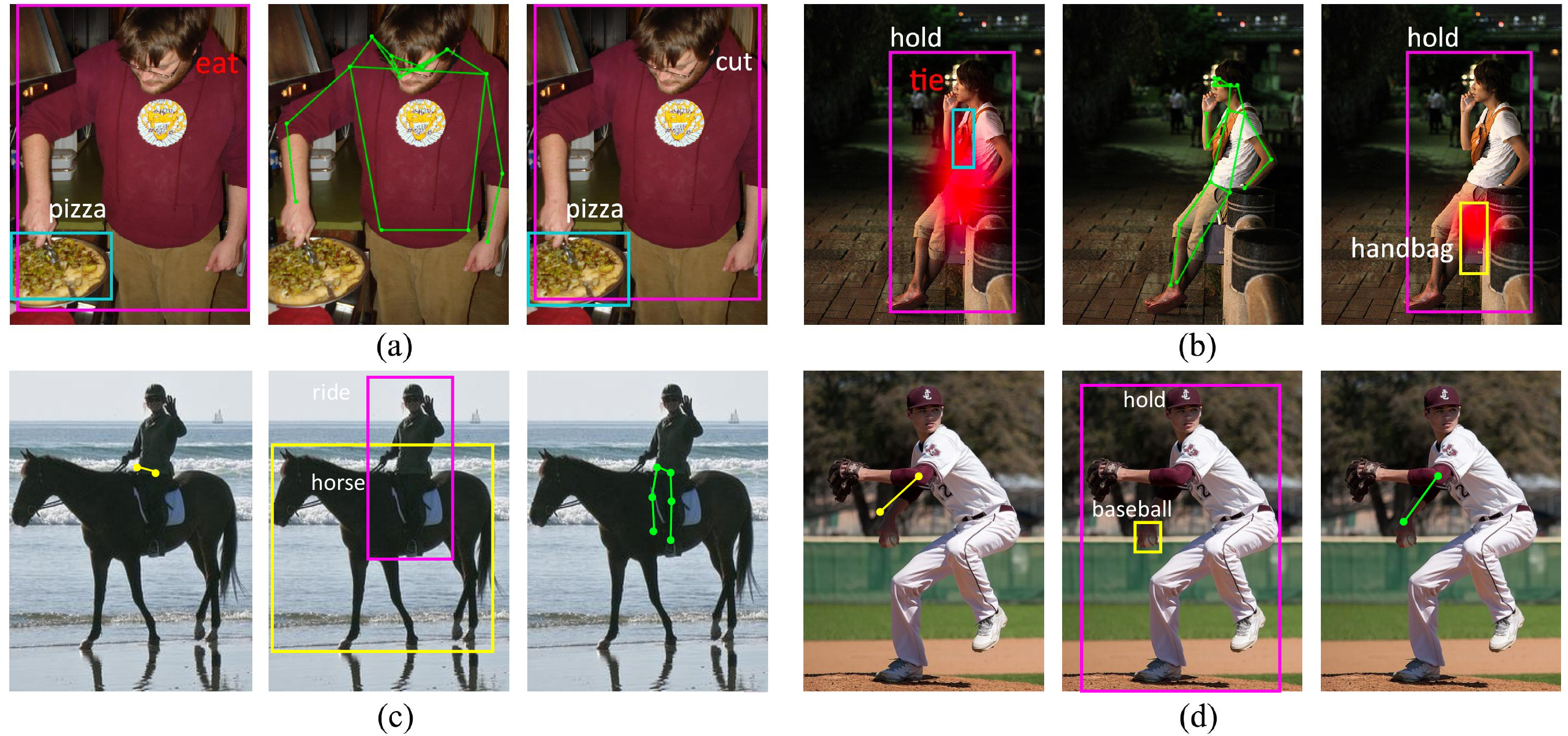}}
	\end{minipage}
	\caption{Examples showing that pose estimation helps HOI  recognition (a, b) and that HOI recognition helps pose estimation (c, d). In each group of figures, from left to right: original recognition results, the other task's recognition results and recognition results with the other task's features. (c,d) only show the keypoints to be improved.}
	\label{fig:wo_pose}
\end{figure*}


\begin{figure*}
	\centering
	\begin{minipage}[t]{1\linewidth}
		\centerline{\includegraphics[width=0.8\textwidth]{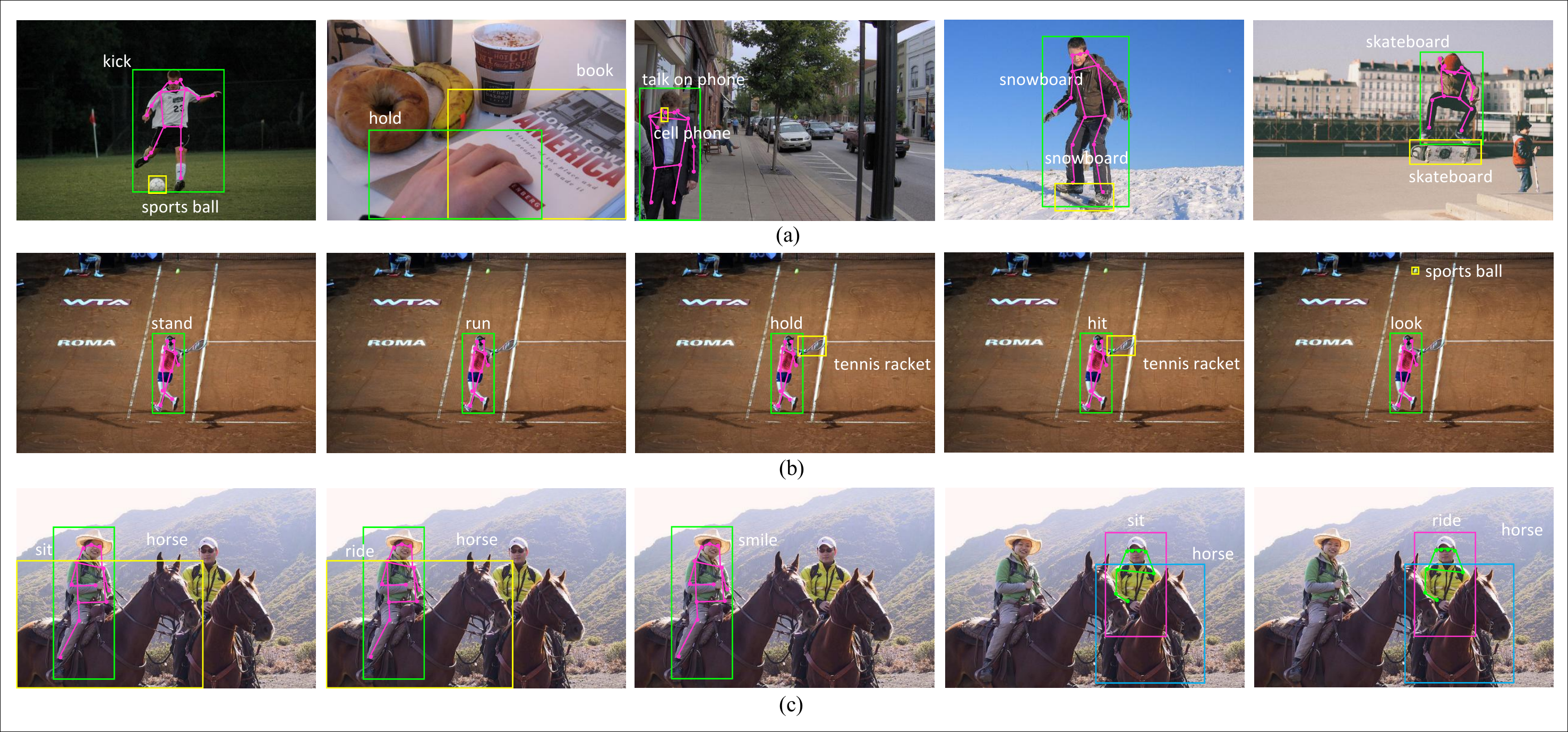}}
	\end{minipage}
	\caption{HOI recognition and pose estimation results. (a) Each image shows one detected $<human, verb, object>$ triplet. (b) One person can take several actions and have several interacted objects. (c) Our method can detect multiple persons who take several actions and have several interacted objects.}
	\label{fig:comparsion}
\end{figure*}

\subsection{Implementation Details}

Our implementation is based on Faster R-CNN~\cite{ren2015faster} built on ResNet-101~\cite{he2016deep}, and the Region Proposal Network (RPN) is frozen and does not share features with our framework for convenient ablation. We extract $7\times7$ features from region proposals by ROIAlign, and the HOI recognition branch consists of two 1024-d fully connected layers followed by dropout layers. The object detection branch consists of 3 residual blocks followed by specific output layers, and the pose estimation branch consists of 8 convolution layers followed by 2 upsample layers(deconv, interp, conv) and output layer(keypoint). The weight decay is 0.0001 and the momentum is 0.9. We use synchronized SGD on 8 GPUs, with each GPU hosting 1 image (the effective mini-batch size per iteration is 8 images). We train the network for 10k iterations with a learning rate of 0.001 and an additional 3k iterations with a learning rate of 0.0001. 

For V-COCO, RPN and the object detection branch are initialized by a Faster R-CNN model pre-trained on MS-COCO, then we train on the V-COCO $trainval$ (5k images) split and report results on the $test$ (5k images) split. We fine-tune the object detection branch and train other branches from scratch. 

The objects in HICO-DET are not exhaustively annotated. We adopt the same measure in~\cite{gkioxari2017detecting}, using a ResNet50-FPN object detector pre-trained on COCO to detect objects in HICO-DET, and the object detection branch is kept frozen during training. Due to the lack of pose annotations in HICO-DET, keypoint detection branch cannot be trained as well. Thus, we implement two strategies to adapt to the situations where only the annotation of HOI is available. The details of these two strategies will be discussed later.

\subsection{Model Performance Analysis}

In order to demonstrate the effectiveness of turbo learning framework, we conduct some comparison experiments on the V-COCO dataset. We also compare two strategies on the HICO-DET dataset, where only the annotation of HOI is available. 

\subsubsection{HOI Recognition with vs. without Pose Estimation}

The pose features are important cues for recognizing actions and localizing the interacted objects. Without pose features, action recognition may be misleading by the human appearance. Meanwhile, the target localization may be confused when there are close objects. To demonstrate the importance of pose estimation to HOI recognition, we evaluate a variant of our method in which only has the HOI recognition model and the number of stages is one. The results are shown in Fig~\ref{fig:HOI}(a) and Fig~\ref{fig:HOI}(b). Removing the pose estimation branch shows a degradation of 0.5\% in $AP_{agent}$ and 1.1\% in $AP_{role}$, which demonstrates the contribution of pose estimation to HOI recognition. To demonstrate the benefits of regarding pose features as input, we also report the results of simple multi-task training in Fig~\ref{fig:HOI}(a) and Fig~\ref{fig:HOI}(b). Multi-task training means that when the number of stages is one, pose estimation branch and HOI recognition branch are jointly learned, but the pose features like $P_{mid}$ and $P_{out}$ do not input to the HOI recognition branch. Using explicit input shows an improvement of 0.2\% in $AP_{agent}$ and 0.8\% in $AP_{role}$, which demonstrates the benefits of explicit input.

We also show some qualitative results in Fig~\ref{fig:wo_pose}(a) and Fig~\ref{fig:wo_pose}(b). In the Fig~\ref{fig:wo_pose}(a), the action is easily classified as eating because of the presence of pizza, but actually the pizza is far from the keypoint of the face. With the pose features, the action is correctly classified as cutting. In the Fig~\ref{fig:wo_pose}(b), the estimated area is in front of the human body. However, the estimated area is near the left wrist with the pose features, which help the model to successfully locate objects.

\subsubsection{Pose Estimation with vs. without HOI Recognition}

To demonstrate that HOI can guide the keypoints distribution, we also evaluate a variant of our method which removes the HOI recognition branch, so the pose estimation task only relies on the image features extracted by the ROIAlign layer. The pose estimation branch is the same as Mask-RCNN~\cite{He2017Mask} architecture, so ``w/o HOI recognition" is actually the result of the Mask-RCNN baseline, and ``RMPE'' is the result of the RMPE~\cite{fang2017rmpe} baseline. But in order to perform HOI recognition, we need not only detect the person, but also detect other kinds of objects. Fig~\ref{fig:HOI}(c) shows that removing HOI recognition task leads to 1\% drop on $AP^{kp}_{50}$, indicating that the HOI recognition features provide guidance on the relative distribution of keypoints. 

We show two examples of improvement in keypoints detection in Fig~\ref{fig:wo_pose}. In the Fig~\ref{fig:wo_pose}(c), the legs are easy to be missed when people ride horse. However, the action ``ride" has similar keypoints distribution. With the HOI recognition features, the keypoints are correctly detected.
In the Fig~\ref{fig:wo_pose}(d), the location of baseball provides guidance  for the keypoints of wrists, which improves the keypoints localization.

\subsubsection{Benefits of Turbo Learning Framework}

Turbo learning framework aims to reuse the cooperation of two tasks, and refine both of the results. To demonstrate the benefits of turbo learning framework, we show the recognition results from stage 1 to stage 3 in Fig~\ref{fig:HOI}. From stage 1 to stage 2, the improvement of $AP_{agent}$, $AP_{role}$ and $AP_{50}^{kp}$ is 1.7\%, 0.2\% and 0.4\% respectively. From stage 2 to stage 3, the improvement of $AP_{agent}$, $AP_{role}$ and $AP_{50}^{kp}$ is 0.9\%, 1.3\% and 0.8\% respectively. As the number of stages increases, the results of both HOI recognition and pose estimation are improved.


\begin{table}   	
	\begin{tabular}{cccc}  
		\hline  
		Method & full  & rare & non-rare \\
		\hline  
		{Gupta $\&$ Malik~\cite{gupta2015visual}} & 7.81  & 5.37 & 8.54 \\
		{InteractNet~\cite{gkioxari2017detecting}} & 9.94 & 7.16  & 10.77 \\
		{Proposed (Pre-train)} & 10.9 & 6.5 & 12.2 \\
	    {Proposed (Semiautomatic)} & \textbf{11.4} & \textbf{7.3} & \textbf{12.6} \\
		\hline  
	\end{tabular}  
	\caption{Results on HICO-DET test set.}  
	\label{HICO}  
\end{table}  

\begin{table}   
	\begin{tabular}{ccc}  
		\hline  
		Method & $AP_{agent}$ & $AP_{role}$ \\
		\hline  
		{Gupta $\&$ Malik~\cite{gupta2015visual}} & 65.1 & 31.8  \\
		{InteractNet~\cite{gkioxari2017detecting}} & 69.2 & 40.0 \\
		{Proposed} & \textbf{70.3} & \textbf{42.0} \\
		\hline  
	\end{tabular}  
	\caption{Results on V-COCO test set.}  
	\label{V-COCO}  
\end{table}

\subsubsection{Pre-training vs. Semiautomatic Annotation}

Although there are not many datasets where both annotations are available, e.g. HICO-DET dataset, we implement two strategies to avoid the lack of pose annotations. The first strategy is pre-training, in which we pre-train the model on the dataset which has both annotations. Then we finetune the network from the model pre-trained on V-COCO and remove the keypoint loss. Still, we allow the weights in keypoint detection branch to be updated following gradients produced by HOI recognition loss, and we find it yields better results. The second strategy is semiautomatic annotation, in which we use the model proposed in~\cite{newell2017associative} to semi-automatically annotate the keypoints on HICO-DET dataset. The used model is trained on the COCO dataset, which is readily applicable
to other open source pose detectors like Mask-RCNN. Then we take the keypoints detection results as annotations for training the model on HICO-DET dataset.   

In Table~\ref{HICO}, we show the comparison of two strategies. The experiment on HICO-DET in~\cite{gkioxari2017detecting} has the similar settings as ours, except that we do not use the Feature Pyramid Network (FPN)~\cite{Lin2016Feature} and interaction branch in ~\cite{gkioxari2017detecting}. So we adopt  their results as a solid baseline. Table~\ref{HICO} shows that pre-training on a dataset having both annotations will improve the HOI recognition results, which is mainly because precise human body structure information learned on V-COCO can help the model to be more robust to the changes in human appearance. The last row of Table~\ref{HICO} shows that semiautomatic annotation can further improve the HOI recognition results, which verifies that our model is also applicable to the case where both HOI and keypoint annotations are not simultaneously available.

\subsection{Comparsion with the State-of-the-art Methods}

As shown in Tables~\ref{HICO} and~\ref{V-COCO}, the proposed method can achieve the state-of-the-art performance on HICO-DET and V-COCO. As~\cite{gupta2015visual} only reported $AP_{role}$ on a subset that consists of 19 actions and only evaluated on the val set of V-COCO, we adopt the solid baseline implemented in~\cite{gkioxari2017detecting}. It is worth noting that our method does not use the FPN and interaction branch as~\cite{gkioxari2017detecting}, but the final $AP_{role}$ still outperforms 2$\%$ on V-COCO, which further proves the effectiveness of our method. Although the keypoints of HICO-DET are obtained by semiautomatic annotation rather than ground truth, our method still outperforms 1.5\% than InteractNet.

We also show some HOI recognition and pose estimation results of the proposed model. In Fig~\ref{fig:comparsion}(a), we only show one triplet $<human,action,object>$. The results of one person with several actions are shown in Fig~\ref{fig:comparsion}(b), and results of multiple persons are shown in Fig~\ref{fig:comparsion}(c). These results show our model can handle different HOI cases.

\section{Conclusion}
We propose a turbo learning method to perform both HOI recognition and pose estimation. As these two tasks can provide guidance to each other, we introduce two novel modules: pose aware HOI recognition module and HOI guided pose estimation module, in which each task's features are also treated as a part of input to the other task. These two modules form a closed loop to utilize complementary information iteratively, which are trained end-to-end. As the number of iterations increases, both results are improved gradually. The proposed method has achieved the state-of-the-art performance on V-COCO and HICO-DET datasets.

\section*{Acknowledgement}
This work was supported in part by the National Natural Science Foundation of China under Grant Nos. 61332007 and 61621136008.

\bibliographystyle{aaai}
\bibliography{AAAI-FengWei.1805}

\begin{thebibliography}{}

\bibitem[\protect\citeauthoryear{Cao \bgroup et al\mbox.\egroup
  }{2017}]{cao2017realtime}
Cao, Z.; Simon, T.; Wei, S.-E.; and Sheikh, Y.
\newblock 2017.
\newblock Realtime multi-person 2d pose estimation using part affinity fields.
\newblock In {\em Proceedings of the IEEE Conference on Computer Vision and
  Pattern Recognition}, volume~1, ~7.

\bibitem[\protect\citeauthoryear{Chao \bgroup et al\mbox.\egroup
  }{2015}]{chao2015hico}
Chao, Y.-W.; Wang, Z.; He, Y.; Wang, J.; and Deng, J.
\newblock 2015.
\newblock Hico: A benchmark for recognizing human-object interactions in
  images.
\newblock In {\em Proceedings of the IEEE International Conference on Computer
  Vision},  1017--1025.

\bibitem[\protect\citeauthoryear{Chao \bgroup et al\mbox.\egroup
  }{2017}]{chao2017learning}
Chao, Y.-W.; Liu, Y.; Liu, X.; Zeng, H.; and Deng, J.
\newblock 2017.
\newblock Learning to detect human-object interactions.
\newblock In {\em arXiv preprint arXiv:1702.05448}.

\bibitem[\protect\citeauthoryear{Chen and Grauman}{2014}]{chen2014predicting}
Chen, C.-Y., and Grauman, K.
\newblock 2014.
\newblock Predicting the location of “interactees” in novel human-object
  interactions.
\newblock In {\em Proceedings of the Asian conference on computer vision},
  351--367.
\newblock Springer.

\bibitem[\protect\citeauthoryear{Delaitre \bgroup et al\mbox.\egroup
  }{2010}]{Delaitre2010Recognizing}
Delaitre, V.; Laptev, I.; and Sivic, J.
\newblock 2010.
\newblock Recognizing human actions in still images: a study of bag-of-features
  and part-based representations.
\newblock In {\em Proceedings of the British Machine Vision Conference},
  1--11.

\bibitem[\protect\citeauthoryear{Desai and Ramanan}{2012}]{Desai2012Detecting}
Desai, C., and Ramanan, D.
\newblock 2012.
\newblock Detecting actions, poses, and objects with relational phraselets.
\newblock In {\em Proceedings of the European Conference on Computer Vision},
  158--172.
\newblock Springer.

\bibitem[\protect\citeauthoryear{Fang \bgroup et al\mbox.\egroup
  }{2017}]{fang2017rmpe}
Fang, H.; Xie, S.; Tai, Y.-W.; and Lu, C.
\newblock 2017.
\newblock Rmpe: Regional multi-person pose estimation.
\newblock In {\em Proceedings of the IEEE International Conference on Computer
  Vision}, volume~2.

\bibitem[\protect\citeauthoryear{Gkioxari \bgroup et al\mbox.\egroup
  }{2017}]{gkioxari2017detecting}
Gkioxari, G.; Girshick, R.; Doll{\'a}r, P.; and He, K.
\newblock 2017.
\newblock Detecting and recognizing human-object interactions.
\newblock In {\em arXiv preprint arXiv:1704.07333}.

\bibitem[\protect\citeauthoryear{Gupta and Malik}{2015}]{gupta2015visual}
Gupta, S., and Malik, J.
\newblock 2015.
\newblock Visual semantic role labeling.
\newblock In {\em arXiv preprint arXiv:1505.04474}.

\bibitem[\protect\citeauthoryear{Gupta \bgroup et al\mbox.\egroup
  }{2009}]{gupta2009observing}
Gupta, A.; Kembhavi, A.; and Davis, L.~S.
\newblock 2009.
\newblock Observing human-object interactions: Using spatial and functional
  compatibility for recognition.
\newblock In {\em Proceedings of the IEEE Transactions on Pattern Analysis and
  Machine Intelligence}, volume~31,  1775--1789.
\newblock IEEE.

\bibitem[\protect\citeauthoryear{He \bgroup et al\mbox.\egroup
  }{2016}]{he2016deep}
He, K.; Zhang, X.; Ren, S.; and Sun, J.
\newblock 2016.
\newblock Deep residual learning for image recognition.
\newblock In {\em Proceedings of the IEEE Conference on Computer Vision and
  Pattern Recognition},  770--778.

\bibitem[\protect\citeauthoryear{He \bgroup et al\mbox.\egroup
  }{2017}]{He2017Mask}
He, K.; Gkioxari, G.; Doll{\'a}r, P.; and Girshick, R.
\newblock 2017.
\newblock Mask r-cnn.
\newblock In {\em Proceedings of the IEEE International Conference on Computer
  Vision},  2980--2988.
\newblock IEEE.

\bibitem[\protect\citeauthoryear{Herath \bgroup et al\mbox.\egroup
  }{2017}]{herath2017going}
Herath, S.; Harandi, M.; and Porikli, F.
\newblock 2017.
\newblock Going deeper into action recognition: A survey.
\newblock In {\em Proceedings of the Image and Vision Computing}, volume~60,
  4--21.
\newblock Elsevier.

\bibitem[\protect\citeauthoryear{Hu \bgroup et al\mbox.\egroup
  }{2013}]{Hu2014Recognising}
Hu, J.-F.; Zheng, W.-S.; Lai, J.; Gong, S.; and Xiang, T.
\newblock 2013.
\newblock Recognising human-object interaction via exemplar based modelling.
\newblock In {\em Proceedings of the IEEE International Conference on Computer
  Vision},  3144--3151.
\newblock IEEE.

\bibitem[\protect\citeauthoryear{Lin \bgroup et al\mbox.\egroup
  }{2014}]{lin2014microsoft}
Lin, T.-Y.; Maire, M.; Belongie, S.; Hays, J.; Perona, P.; Ramanan, D.;
  Doll{\'a}r, P.; and Zitnick, C.~L.
\newblock 2014.
\newblock Microsoft coco: Common objects in context.
\newblock In {\em Proceedings of the European Conference on Computer Vision},
  740--755.
\newblock Springer.

\bibitem[\protect\citeauthoryear{Lin \bgroup et al\mbox.\egroup
  }{2017}]{Lin2016Feature}
Lin, T.-Y.; Doll{\'a}r, P.; Girshick, R.; He, K.; Hariharan, B.; and Belongie,
  S.
\newblock 2017.
\newblock Feature pyramid networks for object detection.
\newblock In {\em Proceedings of the IEEE Conference on Computer Vision and
  Pattern Recognition}, ~4.

\bibitem[\protect\citeauthoryear{Liu \bgroup et al\mbox.\egroup
  }{2018}]{liu2018cascaded}
Liu, W.; Chen, J.; Li, C.; Qian, C.; Chu, X.; and Hu, X.
\newblock 2018.
\newblock A cascaded inception of inception network with attention modulated
  feature fusion for human pose estimation.
\newblock In {\em Thirty-Second AAAI Conference on Artificial Intelligence}.

\bibitem[\protect\citeauthoryear{Luvizon \bgroup et al\mbox.\egroup
  }{2017}]{luvizon2017human}
Luvizon, D.~C.; Tabia, H.; and Picard, D.
\newblock 2017.
\newblock Human pose regression by combining indirect part detection and
  contextual information.
\newblock In {\em arXiv preprint arXiv:1710.02322}.

\bibitem[\protect\citeauthoryear{Luvizon \bgroup et al\mbox.\egroup
  }{2018}]{luvizon20182d}
Luvizon, D.~C.; Picard, D.; and Tabia, H.
\newblock 2018.
\newblock 2d/3d pose estimation and action recognition using multitask deep
  learning.
\newblock In {\em arXiv preprint arXiv:1802.09232}.

\bibitem[\protect\citeauthoryear{Maji \bgroup et al\mbox.\egroup
  }{2011}]{maji2011Action}
Maji, S.; Bourdev, L.; and Malik, J.
\newblock 2011.
\newblock Action recognition from a distributed representation of pose and
  appearance.
\newblock In {\em Proceedings of the IEEE Conference on Computer Vision and
  Pattern Recognition},  3177--3184.
\newblock IEEE.

\bibitem[\protect\citeauthoryear{Mallya and
  Lazebnik}{2016}]{mallya2016learning}
Mallya, A., and Lazebnik, S.
\newblock 2016.
\newblock Learning models for actions and person-object interactions with
  transfer to question answering.
\newblock In {\em Proceedings of the European Conference on Computer Vision},
  414--428.
\newblock Springer.

\bibitem[\protect\citeauthoryear{Newell \bgroup et al\mbox.\egroup
  }{2017}]{newell2017associative}
Newell, A.; Huang, Z.; and Deng, J.
\newblock 2017.
\newblock Associative embedding: End-to-end learning for joint detection and
  grouping.
\newblock In {\em Advances in Neural Information Processing Systems},
  2277--2287.

\bibitem[\protect\citeauthoryear{Ning \bgroup et al\mbox.\egroup
  }{2017}]{ning2017knowledge}
Ning, G.; Zhang, Z.; and He, Z.
\newblock 2017.
\newblock Knowledge-guided deep fractal neural networks for human pose
  estimation.
\newblock In {\em Proceedings of the IEEE Transactions on Multimedia}.
\newblock IEEE.

\bibitem[\protect\citeauthoryear{Pishchulin \bgroup et al\mbox.\egroup
  }{2014}]{pishchulin2014fine}
Pishchulin, L.; Andriluka, M.; and Schiele, B.
\newblock 2014.
\newblock Fine-grained activity recognition with holistic and pose based
  features.
\newblock In {\em Proceedings of the German Conference on Pattern Recognition},
   678--689.
\newblock Springer.

\bibitem[\protect\citeauthoryear{Ren \bgroup et al\mbox.\egroup
  }{2015}]{ren2015faster}
Ren, S.; He, K.; Girshick, R.; and Sun, J.
\newblock 2015.
\newblock Faster r-cnn: Towards real-time object detection with region proposal
  networks.
\newblock In {\em Proceedings of the Advances in Neural Information Processing
  Systems},  91--99.

\bibitem[\protect\citeauthoryear{Shen \bgroup et al\mbox.\egroup
  }{2018}]{shen2018scaling}
Shen, L.; Yeung, S.; Hoffman, J.; Mori, G.; and Fei-Fei, L.
\newblock 2018.
\newblock Scaling human-object interaction recognition through zero-shot
  learning.
\newblock In {\em 2018 IEEE Winter Conference on Applications of Computer
  Vision},  1568--1576.
\newblock IEEE.

\bibitem[\protect\citeauthoryear{Wei \bgroup et al\mbox.\egroup
  }{2016}]{wei2016convolutional}
Wei, S.-E.; Ramakrishna, V.; Kanade, T.; and Sheikh, Y.
\newblock 2016.
\newblock Convolutional pose machines.
\newblock In {\em Proceedings of the IEEE Conference on Computer Vision and
  Pattern Recognition},  4724--4732.

\bibitem[\protect\citeauthoryear{Xiaohan~Nie \bgroup et al\mbox.\egroup
  }{2015}]{xiaohan2015joint}
Xiaohan~Nie, B.; Xiong, C.; and Zhu, S.-C.
\newblock 2015.
\newblock Joint action recognition and pose estimation from video.
\newblock In {\em Proceedings of the IEEE Conference on Computer Vision and
  Pattern Recognition},  1293--1301.

\bibitem[\protect\citeauthoryear{Yao and Fei-Fei}{2010}]{yao2010modeling}
Yao, B., and Fei-Fei, L.
\newblock 2010.
\newblock Modeling mutual context of object and human pose in human-object
  interaction activities.
\newblock In {\em Proceedings of the IEEE Conference on Computer Vision and
  Pattern Recognition},  17--24.
\newblock IEEE.

\end{thebibliography}

\end{document}